\documentclass{article}

\usepackage{microtype}
\usepackage{graphicx}
\usepackage{subfigure}
\usepackage{booktabs} 

\usepackage{subfigure} 
\usepackage{graphicx}
\usepackage{amssymb}
\usepackage{amsmath}
\usepackage{amsthm}
\usepackage{mathtools}
\usepackage{color}
\usepackage{enumitem}
\usepackage{tikz}
\usetikzlibrary{external}
\usetikzlibrary{arrows.meta}
\usetikzlibrary{shapes}
\usetikzlibrary{plotmarks}

\usepackage{hyperref}

\usepackage[accepted]{icml2019}

\DeclareMathOperator*{\argmax}{arg\,max}     
\DeclareMathOperator*{\argmin}{arg\,min}

\DeclarePairedDelimiterX{\infdivx}[2]{[}{]}{%
  #1\;\delimsize\|\;#2%
}

\newcommand{\E}{\mathbb{E}}
\DeclareMathOperator{\bernoullipdf}{Bernoulli}

\DeclarePairedDelimiter\abs{\lvert}{\rvert}

\icmltitlerunning{Active Learning for Decision-Making}

\begin{document}

\twocolumn[
\icmltitle{Active Learning for Decision-Making from Imbalanced Observational Data}



\icmlsetsymbol{equal}{*}

\begin{icmlauthorlist}
\icmlauthor{Iiris Sundin}{aalto}
\icmlauthor{Peter Schulam}{equal,jhu}
\icmlauthor{Eero Siivola}{equal,aalto}
\icmlauthor{Aki Vehtari}{aalto}
\icmlauthor{Suchi Saria}{jhu}
\icmlauthor{Samuel Kaski}{aalto}
\end{icmlauthorlist}

\icmlaffiliation{aalto}{Department of Computer Science, Aalto University, Espoo, Finland}
\icmlaffiliation{jhu}{Department of Computer Science, Johns Hopkins University, Baltimore, MD 21218, USA}

\icmlcorrespondingauthor{Iiris Sundin}{iiris.sundin@aalto.fi}

\icmlkeywords{Active learning, decision-support, observational data, causal inference, individual treatment effects, Bayesian methods}

\vskip 0.3in
]



\printAffiliationsAndNotice{\icmlEqualContribution} 

\begin{abstract}
Machine learning can help personalized decision support by learning models to predict individual treatment effects (ITE). This work studies the reliability of prediction-based decision-making in a task of deciding which action $a$ to take for a target unit after observing its covariates $\tilde{x}$ and predicted outcomes $\hat{p}(\tilde{y} \mid \tilde{x}, a)$. An example case is personalized medicine and the decision of which treatment to give to a patient. A common problem when learning these models from observational data is imbalance, that is, difference in treated/control covariate distributions, which is known to increase the upper bound of the expected ITE estimation error. We propose to assess the decision-making reliability by estimating the ITE model's Type S error rate, which is the probability of the model inferring the sign of the treatment effect wrong. Furthermore, we use the estimated reliability as a criterion for active learning, in order to collect new (possibly expensive) observations, instead of making a forced choice based on unreliable predictions. We demonstrate the effectiveness of this decision-making aware active learning in two decision-making tasks: in simulated data with binary outcomes and in a medical dataset with synthetic and continuous treatment outcomes.
\end{abstract}

\section{Introduction}
A promising application domain of machine learning is to augment human intelligence in decision-making tasks by providing predictions of outcomes under alternative actions~\citep{schulam2017reliable}. To fit a model to this task, we need data recording previous actions $a$, observed outcomes $y$, and any features relevant to the context of the decision, $x$. Then, the goal is to estimate $p(Y \mid X=x,A=a)$ and, further, individual treatment effect (ITE), $\tau(x) = \E[Y[1]-Y[0] \mid X=x]$, where $Y[a]$ denotes the potential outcome of treatment $A=a$~\citep{rubin1978bayesian}. ITE provides sufficient information to choose between two actions. The estimation of ITE is susceptible to many error sources~\citep{pearl2009causality, schulam2017reliable, mitchell2018prediction}, of which we concentrate on imbalance~\citep{gelman2007data}.

Imbalance is defined as the difference in covariate distributions in the treated and control groups. Imbalance makes correct model specification essential for avoiding bias in treatment effect estimates~\citep{gelman2007data}. Mis-specification of the model could be avoided by using non-parametric models, but their variance increases quickly when extrapolating. Recently, imbalance has been shown to increase the upper bound of the model error in estimation of ITE~\citep{shalit2017estimating}. Furthermore, imbalance becomes the more prevalent issue the higher-dimensional the covariate space is~\citep{damour2018overlap}.

There are many existing ways to deal with imbalance when learning the average treatment effect (ATE). In causal inference, the most common methods are propensity score matching or weighting~\citep{rosenbaum1984reducing,hirano2003efficient,lunceford2004stratification}, and modeling the potential outcomes~\citep{imbens2015causal,hernan2018causal}, as well as doubly robust methods which implement both~\citep{bang2005doubly,funk2011doubly}.

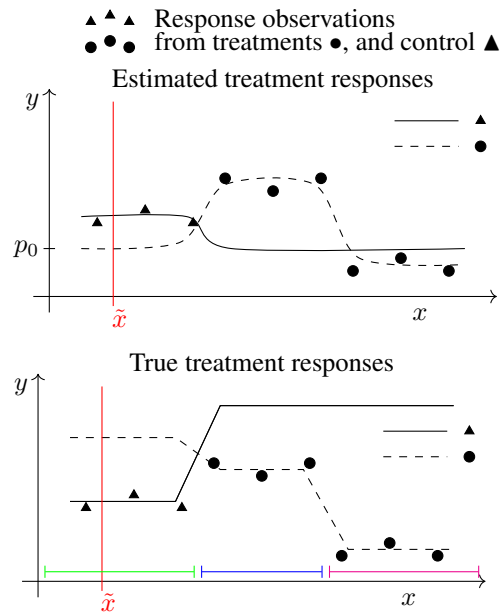
\begin{figure}[bt!]
\centering
\setlength{\tabcolsep}{1pt}
\begin{tabular}[c]{ll}
\centering
~&
\begin{tikzpicture}[scale=0.85]
\draw[->] (-0.25,0.25)--(7,0.25);
\node[anchor= north] at (5.8,0.2) {$x$};
\draw[->] (0,-0)--(0,3.5);
\node[anchor= east] at (-0.01, 3.3) {$y$};
\draw [dashed] plot [smooth] coordinates {(0.5,1.0)  (2.05,1.1)  (2.75,2.0)  (4.15,2)  (4.85,0.85)  (6.5,0.75)};
\draw plot [smooth] coordinates {(0.5,1.5) (2.15,1.5)  (2.85,1.0)   (6.5,1.0)};
\foreach \Point in {(0.75, 1.4), (1.5,1.6)}{
    \node at \Point {\textcolor{black}{\pgfuseplotmark{triangle*}}};
}
\node at (2.25, 1.4) {\pgfuseplotmark{triangle*}};
\foreach \Point in {(2.75, 2.1), (3.5,1.9), (4.25,2.1), (4.75, 0.65), (5.5, 0.85), (6.25,0.65)}{
    \node at \Point {\pgfuseplotmark{*}};
}
\node at (0.65,4.55) {\pgfuseplotmark{triangle*}};
\node at (0.95,4.65) {\pgfuseplotmark{triangle*}};
\node at (1.25,4.55) {\pgfuseplotmark{triangle*}};
\node at (0.65,4.15) {\pgfuseplotmark{*}};
\node at (0.95,4.25) {\pgfuseplotmark{*}};
\node at (1.25,4.15) {\pgfuseplotmark{*}};
\node[anchor= west] at (1.5,4.4) [align=left]{Response observations \\ from treatments $\bullet$, and control $\blacktriangle$};
\node[anchor= south] at (3.5,3.3) [align=left]{Estimated treatment responses};
\draw (5.4,3.0) -- (6.4,3.0);
\draw[dashed] (5.4,2.6) -- (6.4,2.6);
\node[anchor= west] at (6.6,3.0) {\pgfuseplotmark{triangle*}};
\node[anchor= west] at (6.6,2.6) {\pgfuseplotmark{*}};
\draw (-0.09,1.0) -- (0.09,1.0);
\node[anchor= east] at (-0.01,1.0) [align=right]{$p_0$};
\draw [red] (1.0,0.1) -- ( 1.0,3.3);
\node[anchor= north] at (1.1,0.2) {\textcolor{red}{$\tilde{x}$}};
\end{tikzpicture} \\
~&
\begin{tikzpicture}[scale=0.85]
\draw[->] (-0.25,0.25)--(7,0.25);
\node[anchor= north] at (5.8,0.2) {$x$};
\draw[->] (0,-0)--(0,3.5);
\node[anchor= east] at (-0.01, 3.3) {$y$};
\draw[dashed] (0.5,2.5) -- (2.15,2.5) -- (2.85,2) -- (4.15,2) -- (4.85,0.75) -- (6.5,0.75) ;
\draw (0.5,1.5) -- (2.15,1.5) -- (2.85,3) -- (6.5,3);
\foreach \Point in {(0.75, 1.4), (1.5,1.6), (2.25, 1.4)}{
    \node at \Point {\pgfuseplotmark{triangle*}};
}
\foreach \Point in {(2.75, 2.1), (3.5,1.9), (4.25,2.1), (4.75, 0.65), (5.5, 0.85), (6.25,0.65)}{
    \node at \Point {\pgfuseplotmark{*}};
}
\node[anchor= south] at (3.5,3.3) [align=left]{True treatment responses};
\draw (5.4,2.6) -- (6.4,2.6);
\draw[dashed] (5.4,2.2) -- (6.4,2.2);
\node[anchor= west] at (6.6,2.6) {\pgfuseplotmark{triangle*}};
\node[anchor= west] at (6.6,2.2) {\pgfuseplotmark{*}};
\draw (0.5,1.5) -- (2.15,1.5) -- (2.85,3) -- (6.5,3);
\draw [red] (1.0,0.1) -- ( 1.0,3.3);
\node[anchor= north] at (1.1,0.2) {\textcolor{red}{$\tilde{x}$}};
\draw[|-|] [green] (0.1, 0.4) -- (2.45, 0.4);
\draw[|-|] [blue] (2.55, 0.4) -- (4.45, 0.4);
\draw[|-|] [magenta] (4.55, 0.4) -- (6.9, 0.4);
\end{tikzpicture}
\end{tabular}
\caption{An example decision-making task is to choose a treatment for a specific $\tilde{x}$ (red mark). Upper graph shows the posterior means of the potential outcome models given observations, and the prior mean $p_0$. The true responses are in the lower graph, which shows that there are three regions with different response types (marked in green, blue and magenta). Lower $y$ is better. The problem is that there are no observations about treatment $\bullet$ in the green region, which causes it to appear to be the best choice for $\tilde{x}$, although that is incorrect.
}
\label{fig:idea}
\end{figure}

Even though these methods can decrease bias in treatment effect estimates, they will increase variance, and therefore may make the decision-making less reliable. This is especially the case with ITE; For example, in the areas of covariate space where there are more control units than treated, intuition is that the model for the treated outcome either has to generalize from less-representative observations (increasing bias) or extrapolate (increasing variance). Either way, there is higher \textit{uncertainty} about the treated outcome, which makes reliable decision-making difficult. An extreme case of this is illustrated in Fig.~\ref{fig:idea}. A natural question then is, could other data sources be exploited instead of making a forced choice based on insufficient observational data.

This work has three main contributions, which are complementary to each other and can be used independently. First, we describe how imbalance decreases \textit{decision-making performance} by increasing Type S error rate~\citep{gelman2000type}, which is the probability of the model inferring the sign of the treatment effect wrong. Second, we propose a Bayesian \textit{estimate for the Type S error rate}, which allows quantifying the reliability of a decision-support model. Third, we propose to alleviate the consequences of imbalance by actively collecting more data. To this end, we introduce \textit{decision-making aware active learning criteria} that improve decision-making performance by minimizing the estimated Type S error rate.

Finally, in many cases there are restrictions on what can be measured. For example, in medicine it is in general not ethical to do an experiment on a patient in order to get information to improve the treatment plan of another patient. Therefore, regular active learning would not be possible and, instead, any new information has to be acquired indirectly. For this reason, we introduce the idea of \textit{counterfactual elicitation} which means soliciting indirect observations about counterfactual outcomes, that is, what would have been the outcome had $x$ been treated with $a'$ instead of $a$.  We demonstrate the effectiveness of the proposed active learning criteria applied to counterfactual elicitation.

Main claims of this paper: 1. Reliability of decision-making is hindered by imbalance in data. 2. Bayesian estimate of the probability of error is a good estimate for the decision-making performance. 3. If there is a way to acquire more data (a simulator, new experiment, ask an expert), \textit{decision-making error can be minimized by decision-making aware active learning}.

Technical contributions
\begin{enumerate}
    \item We state sufficient conditions under which imbalance will cause decision-making error, measured in Type S error rate. 
    \item We introduce the principle of decision-making aware active learning, and propose a decision-making aware acquisition function.
    Our formulation allows both continuous and discrete outcomes.
    \item We propose two types of queries for counterfactual elicitation.
    (a) Observation is a scalar-valued point estimate of a counterfactual outcome from a noisy oracle. (b) Observation is a (potentially erroneous) pairwise comparison between factual and counterfactual outcomes for one unit in the training data.
    \item We show empirically that the proposed estimate of the Type S error rate has strong correlation with the observed Type S error rate, and that the active learning that aims at minimizing the Type S error rate increases the decision-making performance faster than standard active learning methods.
\end{enumerate}

\section{Related Work}
Active learning is commonly used to acquire class labels during model training to improve classification performance, by selecting unlabeled training instances for humans to label. The selection criteria are usually based on uncertainty and correlations of the training instances, see e.g. \citet{fu2013survey} for a survey. Some active learning works also consider richer input than just labels, for example about importance of features \citep{brooks2015featureinsight, ribeiro2016should}. Active learning proposed by \citet{javdani2014near} aims at improving automated decision-making by reducing model uncertainty so that the remaining hypotheses are confined to the same decision region. The importance of active learning for decision-making tasks has been noted in other fields, where e.g. \citet{saar-tsechansky2007decision} developed a heuristic method for deciding which consumers to target in marketing campaigns.

Active learning has also been used to design interventions that improve identifiability of causal networks \citep{hauser2014two}. In addition, \citet{bottou2013counterfactual} studies carefully how counterfactual inference can be used for active learning. Closest to our work is the work on active learning with logged data~\citep{yan2018active}, which proposes a de-biasing query strategy for a classification task. The difference to our work is that they assume the propensities (probability of revealing the label) to be known.

Using data to estimate the effect of interventions has been extensively studied in the field of causal analysis (see e.g. \citet{pearl2009causality,morgan2014counterfactuals,imbens2015causal,hernan2018causal}). In one school of thought, analysts construct a ``causal directed acyclic graph'' embodying substantive knowledge about the domain, and predict the effect of interventions using operations on the graph \citep{pearl2009causality,morgan2014counterfactuals}. Another common approach is to model interventions using \textit{potential outcomes}, where separate random variables are constructed to represent the target outcome under each possible action \citep{neyman1923applications,rubin1978bayesian,neyman1990application,imbens2015causal,hernan2018causal}. Our work builds on recent research on using ideas from causal analysis to learn individual treatment effects. The issue of imbalance is discussed by \citet{johansson2016learning}, and they propose an approach based on empirical risk minimization and domain shift to improve predictions. \citet{xu2016mlhc} estimate the individual treatment effects in a longitudinal setting where individual-specific treatment parameters are refined over time as more observations are collected. \citet{alaa2017bayesian} use Gaussian processes to model individual-specific outcomes under alternative treatments and prove minimax rates on the risk that the approach achieves. Our work builds on these ideas, and is most closely related to the works by \citet{xu2016mlhc} and \citet{alaa2017bayesian}. Although we also use Gaussian processes to model treatment effects, our work is unique in that it leverages the probabilistic framework to design an active learning algorithm to improve the decision-making process.

\section{Problem Formulation}
\label{sec:problem}
\subsection{Setup}
Let $p(y[a] \mid \mathbf{x})$ be the distribution, and $\hat{p}(y[a] \mid \mathbf{x})$ our probabilistic model of the the potential outcome $Y[a]\in \mathbb{R}$ of an action $a\in \{0,1\}$, given covariates $\mathbf{x} \in \mathcal{X}$, e.g. $\mathcal{X}=\mathbb{R}^d$. The decision-maker has a \textit{policy} for choosing which action to take for a unit $\mathbf{x}$, based on its predicted \textit{individual potential outcomes} $\hat{p}(y[a] \mid \mathbf{x})$. The outcome model has been learned in retrospective from observational data, which may be \textit{imbalanced} in the observed actions. The observational data $D$ is a set of $n$ observations $\{y_i, a_i, \mathbf{x}_i\}_{i=1}^{n}$, where $y_i$ is the observed outcome of action $a_i$ for unit $\mathbf{x}_i$. Imbalance means that the covariate distributions are different in the treated ($a_i=1$) and control groups ($a_i=0$).

The \textit{active learning task} is to sequentially improve the decision-making performance under the decision-maker's policy, by making queries about counterfactual outcomes $Y[a]\mid X=\mathbf{x}$. In this work, we use the objective to improve the decisions for a particular target unit $\tilde{\mathbf{x}}$, but the proposed method applies to multiple targets as well.

\subsection{Assumptions for Causal Inference}
We assume that the unknown policy used to choose actions in the training data only depends on the observed covariates $\mathbf{x}\in\mathcal{X}$. This is equivalent to the no unmeasured confounders assumption \citep{hernan2018causal} and implies that all confounders are included in $\mathbf{x}$.

We further assume consistency of potential outcomes, which means that the potential outcomes $p(y[a]\mid \mathbf{x})=p(y\mid X=\mathbf{x}, A=a)$ can be directly estimated from the observed outcomes in the training data. Regardless of this, imbalance will still cause issues.

\section{Methods}

\paragraph{Preliminaries.}
Causal effect of a treatment is the difference between outcomes when a unit $\mathbf{x} \in \mathcal{X}$ is treated and not treated, where $\mathcal{X}$ is the population. Individualized treatment effect is defined as $\tau(\mathbf{x}) = \E[Y[1]-Y[0] \mid X=\mathbf{x}]$, where $a=1$ means treated and $a=0$ not treated, i.e. control. Fundamental problem in causal inference is that we cannot observe both potential outcomes for the same unit $\mathbf{x}$. There exists extensive work on how to learn estimates of the individualized treatment effects $\hat{\tau}(\mathbf{x})$ despite this limitation, e.g.~\citet{hill2011bayesian,johansson2016learning,shalit2017estimating,alaa2017bayesian,wager2018estimation}.

In this work, the task is to decide which action $a$ to choose for $\tilde{\mathbf{x}}$. Decision-making performance is measured as the probability of correct decision, or equivalently, the proportion of correct decisions in repeated decision-making tasks.

\textbf{Definition}: Type S error rate $\gamma$ is the probability of the model inferring the sign of the treatment effect wrong; \\
$\gamma$: $\mathcal{M} \times P_{\mathcal{X},\mathcal{Y}} \rightarrow [0,1]$ where $\mathcal{M}$ is the model space and $P_{\mathcal{X},\mathcal{Y}}$ is the true treatment effects (distributions over $\mathcal{X}\times \mathcal{Y}$). The expected Type S error rate in $\mathcal{X}$ is \\
$\gamma = \E_{P_{\mathcal{X},\mathcal{Y}}}[\mathbb{I}(\text{sign}(\hat{\tau}) \neq \text{sign}(\tau))]$, where $\mathbb{I}(A)=1$ if condition $A$ is true, and $0$ otherwise.

The expected proportion of correct decisions in population $\mathcal{X}$ is then $1-\gamma$, which makes Type S error rate a natural measure of the decision-making performance.

We assume that the decision-maker's policy is to choose the action with the highest expected utility of the outcome. Without loss of generality we will assume that the utility is directly the outcome $y$ (higher better); more sophisticated utilities are discussed in Section~\ref{sec:discussion}.

\subsection{Effect of Imbalance on Type S Error Rate}
In this section we prove that, under certain assumptions, imbalance increases the error rate in decision-making. We start by a sketch of the proof and then continue with details.

\textit{Sketch of proof:} First, we assume a probabilistic model of potential outcomes, with broad prior distributions. This implies that when the sample size is small, posteriors will be wide. Then, we show that imbalance decreases the expected number of samples locally, therefore increasing the Type S error rate locally. Finally, we provide conditions under which local increase in Type S error rate also increases the expected global Type S error rate.

\textbf{Assumption 1. (Prior).} Assume a broad prior on the expected potential outcomes $\mu_a$: $p(\mu_a) > D > 0$ $\forall \mu_a\in[-K,K]$.

\textbf{Assumption 2. (Likelihood).} Likelihood of observation \\
$p(y_a \mid \mu_a) > C > 0$ $\forall y_a \in [-K,K]$.

\textit{Comment.} Consequence of Assumptions 1 and 2 is that if sample size is small, the posterior will be wide.

\textbf{Lemma 1.} Given observations on two potential outcomes $\mathcal{D}=\{y_{1,i}\}_{i=1}^{n_1} \cup \{y_{0,j}\}_{j=1}^{n_0}$, probability of Type S error has lower bound $p(\text{``Type S error''}) > 2K^2 D^2 C^{n_1+n_0}$.

\textbf{Assumption 3. (Covariate distributions).} Let 
$p^a(x):=p(x \mid a)$ be the covariate distribution of group $a$ in covariate space $\mathcal{X}$. Assume $p^a(x)$ are Lipschitz continuous with constant $L$.

\textbf{Definition: (Imbalance).} Imbalance can be measured using Integral Probability Metric as described by \citet{shalit2017estimating}. Let $G$ be a function family consisting of functions $g:\mathcal{X}\rightarrow \mathbb{R}$. For a pair of distributions $p$, $q$ over $\mathcal{X}$ the Integral Probability Metric is defined as
\begin{align}
\label{eq:IPM}
    IPM_G(p,q)=\sup_{g \in G} \left \lvert \int_{\mathcal{X}} g(x)(p(x)-q(x))dx \right \rvert.
\end{align}

\textbf{Assumption 4. (Imbalance).} Assume there exists non-empty $\Omega = \{x\in\mathcal{X} \mid \abs{p^1(x) - p^0(x)} \geq h\}$ where $h > 0$. (For small enough $h$ this holds if there is any imbalance).

\textbf{Lemma 2.} Let $r$ be the smallest radius $r'>0$ of a neighborhood $B_{r'}(x_e)$ of $x_e\in \Omega$, such that $\abs{p^1(x)-p^0(x)}=0$ for some $x$ in the border $\partial B_{r}(x_e)$. Given assumptions 3 and 4, then $r \geq \frac{h}{2L}$ for all $x_e\in\Omega$.

\textbf{Corollary of Lemma 2.} Mark $P_{B_r(x_e)}^a = \int_{B_r(x_e)}p^a(x)dx$. By Lemma 2 and Assumption 3, $P_{B_r(x_e)}^a > P_{B_r(x_e)}^{1-a}$. This means that the expected number of observations from the group $1-a$ is lower than from the group $a$ in $B_r(x_e)$.

The following theorem is our main result:

\textbf{Theorem 1.} Let $N$ be the sample size, and $a$ the treatment with the higher number of observations in $B_r(x_e)$, and $x\in \mathbb{R}$. Then the expected probability of Type S error in $B_r(x_e)$ has lower bound \\
$p(\text{``Type S error''}) > 2K^2 D^2 C^{N (P_{B_r(x_e)}^a - (1-p(a))\frac{h^2}{2L})}$.

\textit{Comment.} Theorem 1 shows that, with fixed $r$, $N$ and $p(a)$, the larger the local imbalance ($h$) in $B_r(x_e)$, the higher the Type S error rate in $B_r(x_e)$ is. Higher-dimensional cases are considered in the supplementary.

Now, we have shown that imbalance increases locally the Type S error rate. Then the question remains whether the error rate increases globally as well, or do the local effects cancel out each other. We prove this in one-dimensional case, but we see no reason why the proof would not extend to higher dimensions as well. The following assumption and theorem give conditions under which imbalance increases the global Type S error rate.

\textbf{Assumption 5.} Assume the following balanced and imbalanced settings. In the balanced setting, let $p^1(x)=p^0(x)=p(x)$, and $x\in \mathbb{R}$. Without loss of generality we assume that imbalance arises from a shift in $p^0(x)$, s.t. in the imbalanced setting $p^0(x) = p^1(x) - \eta(x)$, where $\eta(x) \in\mathbb{R}$, and $\int \eta(x) dx = 0$.

\textbf{Theorem 2}. Denote $P_{\eta \geq h}=\int_{\mathcal{X}}\mathbb{I}(\eta(x) \geq h)p_t(x)dx$, where $p_t(x)$ is the covariate distribution in the test set. Given Assumption 5, if $P_{\eta \geq h} > C^{N(1-u)h}$, then imbalance $\eta(x)$ increases the lower bound of the expected global Type S error rate in $\mathcal{X}$.

Proofs in the supplementary.

\subsection{Estimated Type S Error Rate}
In the Bayesian sense, the model $\hat{p}(y[a] \mid \mathbf{x}, D)$ captures our current understanding of the problem, and therefore the estimated Type S error rate is $\hat{p}(y[1]<y[0] \mid \mathbf{x}, D)$ if the expected effect is positive, that is, if $\E_{\hat{p}(y[1]\mid \mathbf{x}, D)}[y[1]] > \E_{\hat{p}(y[0]\mid \mathbf{x}, D)}[y[0]]$. Respectively, if the expected effect is negative, then the estimated error rate is $\hat{p}(y[1] > y[0] \mid \mathbf{x}, D)$.

We analyze the properties of the estimated Type S error rate in the model family of linear-parameter regression models, using standard Bayesian linear regression with basis functions. The observation model is $y[a] \mid \mathbf{x} \sim N(w^\top \phi(\mathbf{x},a), \sigma_0^2)$, where $\phi(\mathbf{x},a)$ are the basis functions.

The regression weights have Gaussian prior distribution $w \sim N(0,\alpha I)$. Assuming $\sigma_0$ and $\alpha$ are known, the posterior predictive distributions of potential outcomes $\hat{p}(\tilde{y}[a]\mid \tilde{\mathbf{x}}, D)$ are Gaussian, with mean and variance
\begin{align*}
    \hat{\mu}_a(\tilde{\mathbf{x}}) & = \frac{1}{\sigma_0^2} \phi(\tilde{\mathbf{x}},a)^\top S_a \Phi_a^\top y_a \text{ and}
\end{align*}
\begin{align*}
    \hat{\sigma}_a^2(\tilde{\mathbf{x}}) & = \sigma_0^2 + \phi(\tilde{\mathbf{x}},a)^\top S_a \phi(\tilde{\mathbf{x}},a), \text{where} \\
    S_a^{-1} & = \alpha \text{I} + \frac{1}{\sigma_0^2}\Phi_a^\top \Phi_a, 
\end{align*}
where $\Phi_a$ is a matrix containing the feature vectors $\phi(\mathbf{x}_i,a)$ of the $\mathbf{x}_i$ that were treated with $a$, and similarly $y_a$ is a vector of observed outcomes of $a$.

The treatment effect $\tilde{\tau} \mid \tilde{\mathbf{x}} =\tilde{y}[1]-\tilde{y}[0] \mid \tilde{\mathbf{x}} \sim N(\hat{\mu}_1-\hat{\mu}_0, \hat{\sigma}_{a=1}^2(\tilde{\mathbf{x}}) + \hat{\sigma}_{a=0}^2(\tilde{\mathbf{x}}))$. For simplicity, assume that the expected treatment effect $\E_{\hat{p}(\tilde{\tau} \mid \tilde{\mathbf{x}}, D)} [\tilde{\tau}]$ is positive. Then it can easily be shown that the estimated Type S error rate in a test unit $\tilde{\mathbf{x}}$ is 
\begin{align}
\label{eq:gammahat}
    \hat{\gamma}(\tilde{\mathbf{x}}) & = \text{probit}^{-1} \left(-\frac{\abs{\E_{\hat{p}(\tilde{\tau} \mid \tilde{\mathbf{x}}, D)}[\tilde{\tau}] }}{\text{Var}(\hat{p}(\tilde{\tau} \mid \tilde{\mathbf{x}}, D))^{\frac{1}{2}}}\right),
\end{align}
where $\text{probit}^{-1}$ is the cumulative distribution function of normal distribution, and the expectations and variances are over the posterior predictive distribution of $\tilde{\tau}\mid \tilde{\mathbf{x}}$. (The absolute value in (\ref{eq:gammahat}) makes it to apply also to negative expected treatment effects.) From (\ref{eq:gammahat}) we see that the estimated Type S error rate will increase if the estimated treatment effect decreases, or if posterior uncertainty (variance) increases. Intuitively this makes sense.

\subsection{Decision-Making Aware Active Learning}
\label{sec:DMaware}
Our hypothesis is that active learning criteria that reduce the estimated Type S error rate will result in higher decision-making performance. We call active learning criteria that reduce the estimated Type S error \textit{decision-making aware}.

A special example of decision-making aware criteria is \textit{targeted expected information gain} introduced by \citet{sundin2018improving}. It selects the next query by maximizing the expected information gain of the posterior $\hat{p}(\tilde{y} \mid \tilde{x})$ using KL-divergence. This criterion is related to entropy minimization, because the expected KL-divergence between the current and updated posteriors can be written as $\E \left[D_{KL}(\hat{p}^*||\hat{p})\right]=\E\left[H(\hat{p}^*,\hat{p})\right]-\E\left[H(\hat{p}^*)\right]$, where $\hat{p}^*$ is the updated posterior, and $H(\hat{p}^*)$ its entropy, and $H(\hat{p}^*,\hat{p})$ is the cross entropy between the current and updated posteriors. Reducing the entropy in the posterior therefore reduces variance in eq. (\ref{eq:gammahat}), which decreases the estimated Type S error rate.

We propose to directly minimize the estimated Type S error rate in eq. (\ref{eq:gammahat}) with active learning. The criterion is to maximize the estimated reliability of a decision at $\tilde{x}$, that is, $1-\hat{\gamma}(\tilde{x})$.

Directly minimizing the error can be interpreted as only exploiting what we already know, which lacks in exploration. There is an easy fix, though, which is to add exploration on the error. The exploration-exploitation trade-off is managed by maximizing the expected information gain on the predictive distribution of Type S error: $\bernoullipdf(\hat{\gamma}(\tilde{x}))$ (technically, its relative entropy). The maximization of the information gain is equivalent to minimizing the posterior entropy and consequently the log-loss~\citep{settles2012active}. 

\subsection{Counterfactual Elicitation}
Assuming it is possible to acquire (noisy) observations about counterfactuals in the training data, we can do more, as discussed in the introduction. Denote by $\mathbb{D}$ the set of training examples $\{x_i,a_i,y_i\}_{i=1}^{n}$ for which the factual outcomes $y_i$ have been observed, and denote by $\mathbb{U}$ the counterfactual examples $\{x_i,1-a_{i}\}_{i=1}^{n}$, for which the outcomes are unknown. Then we can use active learning to construct a set $\mathbb{L}$ that contains the new observations, and $\mathbb{L}$ will be data which would not normally be available. At each iteration, the algorithm selects $\{x^*,a^*\} \in \mathbb{U}$ to solicit a counterfactual outcome $y^*$. After this, $\{x^*,a^*,y^*\}$ is added to $\mathbb{L}$, and removed from $\mathbb{U}$. So the optimization problem at each query iteration $k$ becomes
\begin{align*}
    &x^*,a^* = \argmin_{\{x,a\} \in \mathbb{U}} \E_{\hat{p}(y\mid x,a,\mathbb{D},\mathbb{L})}  \left[ \hat{\gamma}_{k+1}(\tilde{x}) \right] \text{, where} \\
    &\hat{\gamma}_{k+1}(\tilde{x}) =  \hat{p}\left(y[a_{\tilde{x}}] < y[1-a_{\tilde{x}}]\mid \tilde{x}, \mathbb{D}, \mathbb{L}, \{x,a,y\}\right),
\end{align*}
and $a_{\tilde{x}}$ is the treatment with the highest expected outcome for $\tilde{x}$: $a_{\tilde{x}} = \argmax_{a'} \E_{\hat{p}(y[a']\mid \tilde{x}, \mathbb{D},\mathbb{L}, \{x,a,y\})}[y[a']]$.

Because the new observations are assumed to come from a different source than the original data, the model has separate noise parameters for the observation models of $\mathbb{D}$ and $\mathbb{L}$.

\subsection{Comparative Observations about Counterfactuals}
Another possible way in which noisy observations may be available is as comparisons of two counterfactual outcomes. Then, active learning is used to acquire a comparative observation $c\in{0,1}$, which is a comparison between the expected counterfactual outcomes: $c=1$ if $\E \left[ Y[1] \mid X=x_i\right] > \E \left[ Y[0]\mid X=x_i\right]$, else $c=0$. At each iteration, the algorithm selects $\{x^*\} \in \mathbb{U}$ to solicit comparative observation $c^*$. After this, $\{x^*,c^*\}$ is added to $\mathbb{L}$, and removed from $\mathbb{U}$.

\section{Experiments}
\label{experiments}
We run three sets of experiments. First, we show that imbalance correlates with both the estimated and observed Type S error rate in simulated data. Second, we evaluate the performance of the proposed decision-making active learning criterion in simulated and semi-synthetic medical data. Last, we compare the performance to other active learning criteria that are applicable to the decision-making task, assuming a fixed query budget. 

\subsection{The Observed and Estimated Type S Error Rate in Imbalanced Data}
In this section, we show empirically how imbalance affects the reliability of decision-making, and see that the estimated Type S error rate correlates with the observed error rate. The data are generated such that the outcome model contains interaction between $a$ and $x$, and that the treatment effect is either saturating or increasingly increasing.

The outcome model generation is repeated 200 times, and for each outcome model we generate 6 training sets. The training data generation differs in the propensity scores, resulting in different levels of imbalance in the training data sets. Details of the data generation process are in the Supplementary. We measure imbalance using the Maximum Mean Discrepancy (MDD)~\citep{gretton2012kernel}. We model the potential outcomes using two independent Gaussian Processes with squared exponential kernel.

\begin{figure}
    \centering
    \includegraphics[width=0.55\linewidth]{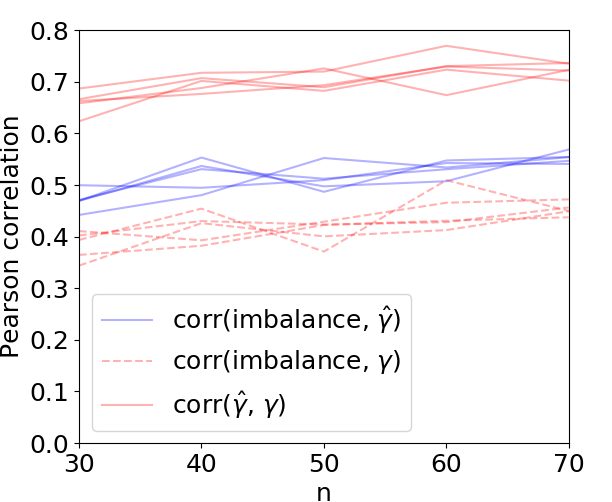}
    \caption{Imbalance correlates with the observed ($\gamma$) and estimated ($\hat{\gamma}$) Type S error rates on wide range of sample sizes $n$. Each line shows correlations between imbalance, estimated Type S error rate and observed Type S error rate in 1200 data sets. Separate lines show variation in 5 repeated experiments.}
    \label{fig:corr_n_per_p}
\end{figure}

Fig.~\ref{fig:corr_n_per_p} shows moderate correlation between imbalance and the Type S error rate, and that the correlation is quite constant across a wide range of sample sizes. This is empirical evidence of Theorem 1. Furthermore, the estimated Type S error rate $\hat{\gamma}$ and observed Type S error rate $\gamma$ have strong correlation, around 0.7. Plotting $\hat{\gamma}$ against $\gamma$ shows that low estimated error indicates low observed error (figure in the Supplementary).
This suggests that the estimated Type S error rate is a good indicator of the prospective decision-making performance.

\subsection{Decision-Making Aware Active Learning}
We evaluate the performance of decision-making aware active learning (D-M Aware) using counterfactual elicitation in two cases. The first is synthetic data with high local imbalance and difficult outcome function shapes with interaction between $a$ and $x$, similar to those in Fig.~\ref{fig:idea}. The second is a semi-synthetic medical data set IHDP~\citep{hill2011bayesian}, commonly used in causal inference.

\subsubsection{Simulated Data with Binary Outcomes}
\label{exp:sim}
In this experiment, we study the proposed active learning approach in simulated data. Binary outcome $y$ indicates the occurrence of an adverse effect, and the decision-making task is therefore to choose the treatment that results in a lower probability of the adverse effect. The setting is similar to that in the Fig.~\ref{fig:idea}.

\paragraph{Synthetic data.} The outcome $y\in \{0,1\}$ is Bernoulli distributed with parameter $\theta_{x,a}$, given a one-dimensional covariate $x \in \mathbb{R}$ and treatment $a \in \{0,1\}$. The data are generated from a logistic regression model with interaction between $a$ and 3 radial basis functions (RBF) $\phi(x)$, s.t. $\theta_{x,a} = \text{logit}^{-1}(\mathbf{w}_0^\top\phi(x) + \mathbf{w}_1^\top \phi(x)a)$. Imbalance is induced to the training data by making the better treatment more likely for each $x$. Training sample size is 30. Details, such as the weights of the radial basis functions, are in the Supplementary.

\paragraph{Model and learning.} We model the data with a logistic regression model $p(y\mid x, a)\sim \bernoullipdf(\theta_{x,a})$, where $\theta_{x,a}$ has the same form as in the data generation process. The model is fit using a probabilistic programming language Stan~\citep{pystan21600, stan2017}. We assume that the RBF centers and length-scale are known, so that only $\mathbf{w}_0$ and $\mathbf{w}_1$ need to be learned.

\subsubsection{Medical Semi-Synthetic Data with Continuous Outcomes}
\label{exp:IHDP}
We evaluate decision-making aware active learning approaches in deciding on medical interventions on real medical data with continuous-valued synthetic outcomes.

\paragraph{The IHDP data set.}
We use the Infant Health and Development Program (IHDP) dataset from \citet{hill2011bayesian}, also used e.g. by \citet{shalit2017estimating} and \citet{alaa2017bayesian}, including synthetic outcomes, containing 747 observations of 25 features. Technical details are that we use the harder of the two cases in the paper, the ``non-overlap case,'' and predict the non-linear outcome involving interactions (``Response Surface C''). These data come from a real randomized experiment, and imbalance has been produced by removing a part of the treated population. We evaluate the performance in leave-one-out cross-validation, but in order to make the problem even more realistically hard, for each of the 747 target units we choose randomly 100 observations as training examples.

\paragraph{Model and learning.} We fit separate GPs to the outcomes of each treatment with GPy\footnote{Toolbox available at:
https://sheffieldml.github.io/GPy} (version 1.9.2), and use mixed noise likelihood to learn the noise in the observations acquired by active learning. We use an exponentiated quadratic kernel with a separate length-scale parameter for each variable, and optimize the hyperparameters using marginal likelihood. Details of the priors are in the Supplementary. We use Gauss-Hermite quadrature of order 32 to approximate the expectations in D-M aware, Targeted-IG, and EIG.

\subsubsection{Results: Evaluation of Decision-Making Aware Active Learning}
\begin{figure}%
\centering
\subfigure[]{%
\label{fig:S_min_simulated}%
\includegraphics[width=0.4\textwidth]{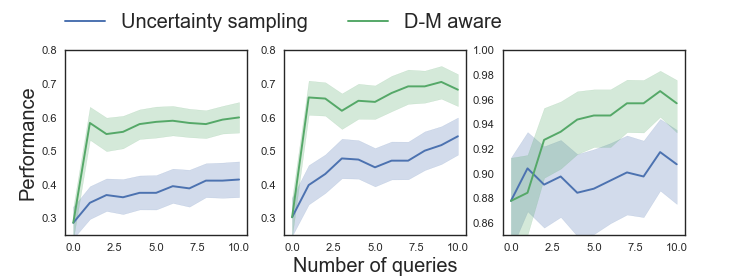}}%
\qquad
\subfigure[]{%
\label{fig:S_min_IHDP}%
\includegraphics[width=0.35\textwidth]{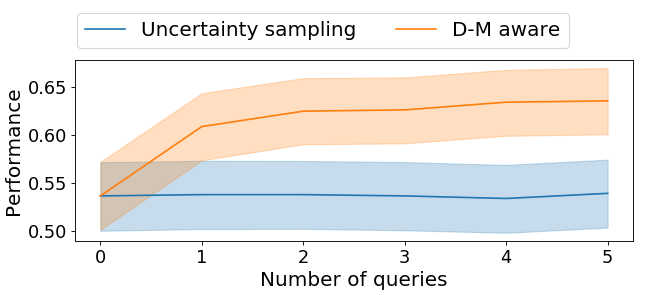}}%
\caption{Active learning that minimizes the estimated Type S error (D-M aware) increases the decision-making performance, measured in proportion of correct decisions. Shaded areas show the 95\% bootstrap confidence interval. (a) Results of the simulated data with binary outcomes, in 100 repetitions averaged over 3 target units in each response type regions; the columns correspond to green (leftmost), blue (middle) and magenta (rightmost) regions in the Fig.~\ref{fig:idea}.  (b) Results in IHDP data as a function of the number of counterfactual queries to a simulated, noisy oracle, averaged over 747 decision-making tasks.}
\label{fig:S_minimization_result}
\end{figure}

First, we evaluate the performance of the proposed active learning that minimizes the estimated Type S error rate (D-M aware) on the decision-making performance, as measured by the proportion of correct decisions. Fig.~\ref{fig:S_minimization_result} shows that our method improves decision-making performance efficiently, compared to the baseline of uncertainty sampling. The difference to the baseline is statistically significant in Fig.~\ref{fig:S_min_simulated} two out of three cases in the simulated data, and Fig.~\ref{fig:S_min_IHDP} in the semi-synthetic medical data (IHDP) (based on 95\% bootstrap confidence intervals).

\subsubsection{Results: Comparison of Active Learning Criteria in Decision-Making Tasks}
Next, we compare the performance of D-M aware to two widely-used earlier active learning approaches, uncertainty sampling and maximum expected information gain \citep{culotta2005reducing, roy2001toward} (EIG), and also include results for the previously introduced special case of decision-making aware active learning, targeted-IG (see Section~\ref{sec:DMaware}). In order to make the methods comparable, we use the variant of D-M aware that also explores (see Section~\ref{sec:DMaware}), as do the EIG and targeted-IG. The performance of the non-exploring variant of D-M aware is slightly lower (not shown) in all cases.

\begin{figure*}%
\centering
\subfigure[]{%
\label{fig:bernoulli_results}%
\includegraphics[width=0.36\textwidth]{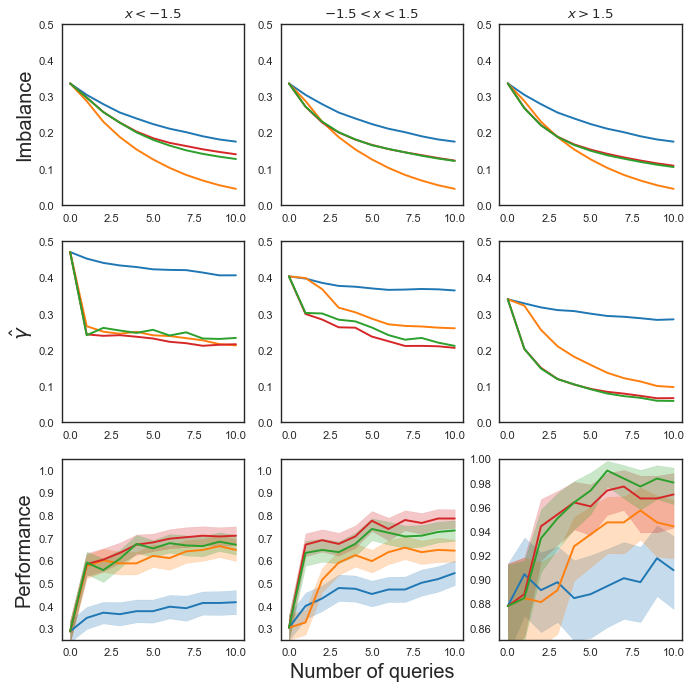}}%
\qquad
\subfigure[]{%
\label{fig:IHDP_decs}%
\includegraphics[width=0.51\textwidth]{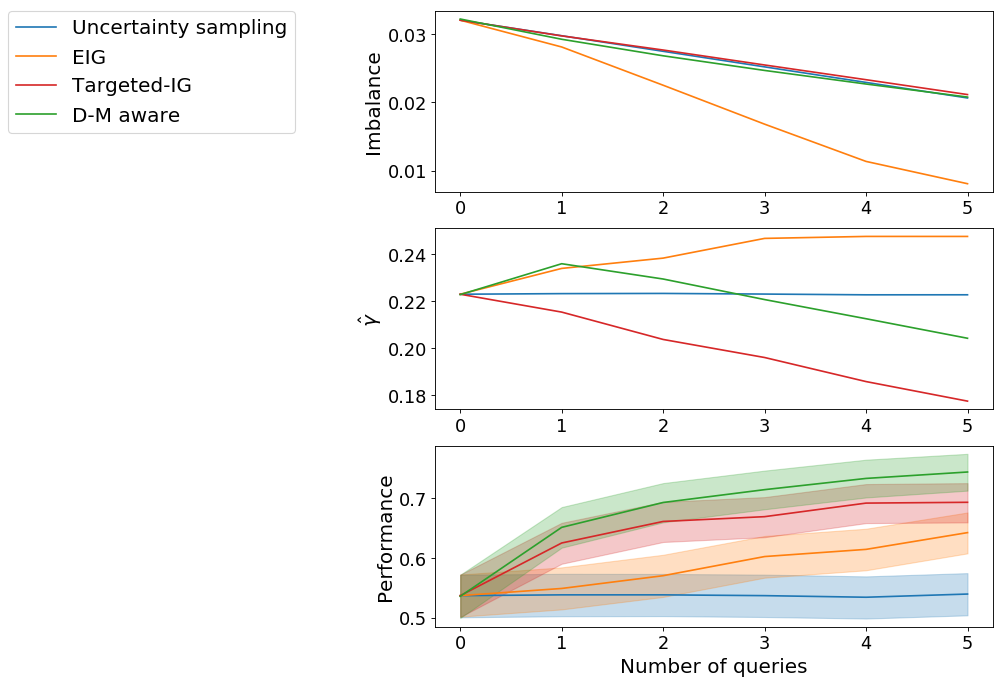}}%
\caption{Comparison of active learning criteria as a function of number of queries. (a) Simulated data (b) IHDP data. The topmost panel in each shows imbalance, middle panel the estimated Type S error rate, and the lowest panel the proportion of correct decisions. Information-gain-based approaches are more effective than uncertainty sampling, and the decision making-aware criteria D-M aware and Targeted-IG are the best. Shaded areas show the 95\% bootstrap confidence interval.
}
\vspace{-0.2cm}
\end{figure*}

\paragraph{Simulated data.} Fig.~\ref{fig:bernoulli_results} shows that the decision-making aware active learning criteria, D-M aware and Targeted-IG, are the fastest to improve decision-making performance, and achieve significant increase in correct decisions after just one query in two cases out of three. After five solicited counterfactuals, Expected Information Gain (EIG) has reached comparable but still lower performance. Uncertainty sampling does not perform well in this example because it concentrates queries to the areas with estimated probability of adverse effect being close to 0.5, instead of close to the target units. Active learning reduces both imbalance and the estimated Type S error rate. Interestingly, D-M aware method achieves good decision-making performance regardless of having only little effect on imbalance, which may be due to local querying that reduces the local imbalance. In contrast, EIG reduces imbalance the most, which is expected as the criterion selects queries that are beneficial to the whole population.

\paragraph{IHDP data.} The results in the IHDP data are similar as in the simulated data; The decision-making performance improves fastest with D-M aware and Targeted-IG, compared to EIG and uncertainty sampling (see Fig.~\ref{fig:IHDP_decs}). D-M aware and Targeted-IG achieve statistically significant improvement in decision-making performance already with one query (based on 95\% bootstrap confidence intervals).

\subsubsection{Comparative Feedback}
\begin{figure}
    \centering
    \includegraphics[width=0.35\textwidth]{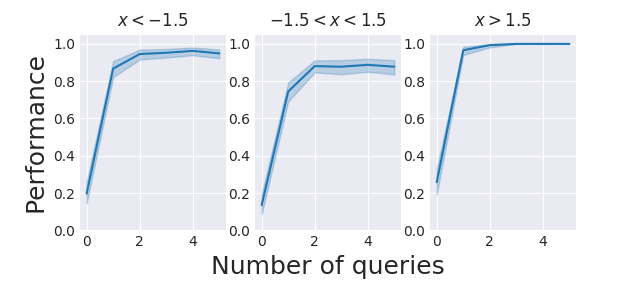}
    \vspace{-0.3cm}
    \caption{Comparative observations on the Bernoulli parameter in the simulated data is effective in increasing the proportion of correct decisions. The results are averaged over 3 target units in each response type regions and 100 repetitions.}
    \vspace{-0.2cm}
    \label{fig:comparative}
\end{figure}

Last, we demonstrate the use of comparative observations for counterfactual elicitation. The setting is the same as in Section~\ref{exp:sim}, with the difference that the query is about which treatment has lower Bernoulli parameter. We fit the outcomes with GPs using Stan~\cite{pystan21600, stan2017}, which allows the model to learn both from direct and comparative observations. The results in Fig.~\ref{fig:comparative} show that comparative observations increase decision-making performance efficiently in the simulated data setting. We note that the results with comparative feedback are better than those in Fig.~\ref{fig:bernoulli_results}, because here the queries give information about $\theta_{x,1}>\theta_{x,0}$, thus providing more information than the direct observations on outcome $y[a]\mid x \sim \bernoullipdf(\theta_{x,a})$. 

\section{Discussion and Conclusions}
\label{sec:discussion}
As machine learning systems are being integrated into human decision-making workflows, it is increasingly important that deployed models are correct and reliable. Predicting likely outcomes of various actions for decision support is especially challenging because the system's success is measured by its ability to correctly forecast the \textit{effect of interventions}. This is more difficult than the classical sense of generalization in machine learning, where the goal is to have low risk under the distribution from which the training data was sampled, meaning that both distributions of $x$ and $a\mid x$ stay the same \citep{pearl2009causality}. In this paper we focused on the effect that imbalance in the training data can have on the reliability of comparisons of $\hat{p}(y[a] \mid \mathbf{x})$. We propose to improve the reliability by active learning that aims at maximizing the estimated reliability. In our experiments, this decision-making aware active learning outperforms standard methods in decreasing the error rate in decision-making.

The most computationally expensive step in the proposed approach is computing the expectation of the information gain, with complexity at worst $N$ times that of re-training the model if no analytic solution is available. For example, a sparse GP model would have $O(N^2M^2)$ complexity where $M$ is the number of inducing points. The complexity of re-training GP models can be reduced using Kalman-filter based implementation with sequential updates for soliciting new observations. Furthermore, the full algorithm can be approximated by only including $k$ nearest neighbours of the target unit $\tilde{x}$, using the model's intrinsic distance measure, when computing the expected utilities. Our preliminary results indicate, however, that the performance suffers if $k$ is too low (see additional results in the Supplementary).

In this work, we assumed that the utility of an action $a$ is directly the outcome $y[a]$. In case the utility is a function of the outcome, $U(y[a])$, our method applies by defining the Type S error rate as the probability of inferring erroneously which utility is the highest.

Future work includes soliciting observations from multiple sources, developing fast approximations that will allow computing the decision-making based utility efficiently for complex models, and studying human experts as one source of information.

One interesting application of the method is in personalized medicine, where counterfactual elicitation could solicit practitioners' knowledge to the model. For example, a clinician in a hospital has access to medical records of previous patients, and may also have personal experience about some of them. These data are rarely included in training sets of the models, but active learning and counterfactual elicitation could allow leveraging this additional source of information to infer more accurately about the future.

\section*{Acknowledgements}
We thank the anonymous reviewers for their comments. This work was supported by the Academy of Finland (grants 298742, 313122, 319264 and 294238), by Alfred Kordelin Foundation and by Jenny and Antti Wihuri Foundation. We acknowledge the computational resources provided by Aalto Science-IT project, and the support from the Finnish Center for Artificial Intelligence (FCAI).

\bibliographystyle{icml2019}
\bibliography{references}

\newpage
\appendix
\section{Effect of Imbalance on Type S Error Rate}
In this section we prove that, under certain assumptions, imbalance increases the error rate in decision-making. We start by a sketch of the proof and then continue with details.

\textit{Sketch of proof:} First, we assume a probabilistic model of potential outcomes, with broad prior distributions. This implies that when the sample size is small, posteriors will be wide. Then, we show that imbalance decreases the expected number of samples locally, therefore increasing the expected Type S error rate locally. Finally, we provide conditions under which local increase in the expected Type S error rate also increases the expected global Type S error rate.

\textbf{Assumption S1. (Prior).} Assume a broad prior on the expected potential outcomes $\mu_a$: $p(\mu_a) > D > 0$ $\forall \mu_a\in[-K,K]$. The action $a\in\{0,1\}$.

\textbf{Assumption S2. (Likelihood).} Likelihood of observation 
$p(y_a \mid \mu_a) > C > 0$ $\forall y_a \in [-K,K]$.

\textit{Comment.} Consequence of Assumptions S1 and S2 is that if sample size is small, the posterior will be wide.

\textbf{Lemma S1.} Given observations on two potential outcomes $\mathcal{D}=\{y_{1,i}\}_{i=1}^{n_1} \cup \{y_{0,j}\}_{j=1}^{n_0}$, the probability of Type S error has lower bound $p(\text{``Type S error''}) > 2K^2 D^2 C^{n_1+n_0}$.

\begin{proof}[Proof:]
We prove Lemma S1 for the case where the true treatment effect is negative, that is, $m_0 > m_1$.

Posterior of $\mu_a$ given data is: \\
$p(\mu_a \mid \mathcal{D}) = \frac{1}{Z}p(y_a \mid \mu_a) p(\mu_a)$.

The probability of the Type S error is
\begin{align}
    p&(\text{``Type S error''}) = \int_{\mu_0 \leq \mu_1 \mid m_0 > m_1} p(\mu_1,\mu_0\mid \mathcal{D}) d\mu_0 d\mu_1 \nonumber 
\end{align}
\begin{align}
    & = \int_{\mu_1}\int_{\mu_0 \leq \mu_1} p(\mu_1\mid \mathcal{D})p(\mu_0\mid \mathcal{D}) d\mu_0 d\mu_1 \label{eq:S1}\\
    & \geq \int_{\mu_1}\int_{\mu_0 \leq \mu_1} \prod_{a=0}^1 \prod_{i=1}^{n_a}p(y_{a,i} \mid \mu_a)p(\mu_a) d\mu_0 d\mu_1 \nonumber\\
    & \geq \int_{-K}^K \int_{-K}^{\mu_1} C^{n_1} D  C^{n_0} D d\mu_0 d\mu_1 \label{eq:S2} \\
    & = D^2 C^{n_1 + n_0} \int_{-K}^K (\mu_1 + K) d\mu_1 \nonumber\\
    & = D^2 C^{n_1 + n_0} 2K^2, \nonumber
\end{align}
where (\ref{eq:S1}) follows from assuming factorization, and (\ref{eq:S2}) by Assumptions S1 and S2.
\end{proof}

\textbf{Assumption S3. (Covariate distributions).} Let 
$p^a(x):=p(x \mid a)$ be the covariate distribution of group $a$ in covariate space $\mathcal{X}$. Assume $p^a(x)$ are Lipschitz continuous with constant $L$.

\textbf{Definition: (Imbalance).} Imbalance can be measured using Integral Probability Metric as described by \citet{shalit2017estimating}. Let $G$ be a function family consisting of functions $g:\mathcal{X}\rightarrow \mathbb{R}$. For a pair of distributions $p$, $q$ over $\mathcal{X}$ the Integral Probability Metric is defined as
\begin{align}
\label{eq:IPM_supp}
    IPM_G(p,q)=\sup_{g \in G} \left \lvert \int_{\mathcal{X}} g(x)(p(x)-q(x))dx \right \rvert.
\end{align}

\textbf{Assumption S4. (Imbalance).} Assume there exists non-empty $\Omega = \{x\in\mathcal{X} \mid \abs{p^1(x) - p^0(x)} \geq h\}$ where $h > 0$. (For small enough $h$ this holds if there is any imbalance).

\textbf{Lemma S2.} Denote $\eta(x) = p^1(x)-p^0(x)$. Then given Assumption S3, $\eta(x)$ is Lipschitz continuous with constant $2L$.

\begin{proof}[Proof:]
\begin{align*}
    \abs{\eta(x) - \eta(x')} & = \abs{p^1(x)-p^0(x) - p^1(x')+ p^0(x')} \\
    & = \abs{p^1(x) - p^1(x') -(p^0(x) - p^0(x'))} \\
    &\leq \abs{p^1(x) - p^1(x')} + \abs{(p^0(x) - p^0(x'))} \\
    &\leq 2L\abs{x-x'}.
\end{align*}
\end{proof}

\textbf{Definition:} Let $r$ be the smallest radius $r'>0$ of a neighborhood $B_{r'}(x_e)$ of $x_e\in \Omega$, such that $\abs{\eta(x)}=0$ for some $x$ in the border $\partial B_{r}(x_e)$.

\textbf{Lemma S3.} Given assumptions S3 and S4, then $r \geq \frac{h}{2L}$ for all $x_e\in\Omega$.

\begin{proof}[Proof:]
Counter-example: Show that if $r < \frac{h}{2L}$, then there does not exist $x\in B_r(x_e)$ for which $\abs{\eta(x)}=0$.

For any $x \in B_r(x_e)$ it holds that
\begin{align}
    \abs{\eta(x_e) - \eta(x)} & \geq \abs{\abs{\eta(x_e)} - \abs{\eta(x)}} \nonumber \\
    & = \abs{\eta(x_e)} - \abs{\eta(x)}  \label{eq:S4}\\
    \Leftrightarrow \abs{\eta(x)} & \geq \abs{\eta(x_e)} - \abs{\eta(x_e) - \eta(x)}  \nonumber \\
    & \geq h - 2L\abs{x_e-x} \label{eq:S5} \\
    & \geq h - 2Lr \nonumber \label{eq:S6} \\
    & > h-2L\frac{h}{2L} = 0,
\end{align}
where the equality in (\ref{eq:S4}) comes from the fact that a necessary condition for $\abs{\eta(x)}=0$ is that $\abs{\eta(x)} < \abs{\eta(x_e)}$, (\ref{eq:S5}) is by Assumption S4, and (\ref{eq:S6}) is due to the counter-assumption $r < \frac{h}{2L}$.

Therefore $\abs{\eta(x)}$ cannot be zero in $\partial B_{r}(x_e)$ unless \\
$r \geq \frac{h}{2L}$.
\end{proof}

\textbf{Lemma S4.} Given assumption S3, $x\in\mathbb{R}$, and assuming $p^a(x_e)>p^{1-a}(x_e)$, then the expected number of samples from the group $1-a$ in $B_r(x_e)$ is upper-bounded by 
$\E[n_{1-a}] \leq (1-p(a))N(P_{B_r(x_e)}^a - \frac{h^2}{2L})$.

\begin{proof}[Proof:]
\begin{align*}
    & \E[n_{1-a}] \\
    & = (1-p(a))N \int_{B_r(x_e)}p^{1-a}(x)dx \\
    & = (1-p(a))N \int_{B_r(x_e)}\left(p^{a}(x) - (p^{a}(x) - p^{1-a}(x))\right)dx \\
    & = (1-p(a))N \left(P_{B_r(x_e)}^a - \int_{B_r(x_e)}(p^{a}(x) - p^{1-a}(x))dx\right) \\
    & = (1-p(a))N \left(P_{B_r(x_e)}^a - \int_{B_r(x_e)}\abs{\eta(x)}dx\right) \\
    & \leq (1-p(a))N \left(P_{B_r(x_e)}^a - \frac{h^2}{2L}\right),
\end{align*}

where we have used $\int_{B_r(x_e)}\abs{\eta(x)}dx \geq \frac{h^2}{2L}$ when $x\in\mathbb{R}$. This comes from the fact that $\abs{\eta(x_e)}\geq h$ (Assumption S4), and by definition $\abs{\eta(x)}=0$ for some $x\in \partial B_r(x_e)$. Thus the integral has its smallest value when $\abs{\eta(x)}$ decreases from $h$ as fast as possible, that is, by Lipschitz constant $2L$ (Lemma S2), s.t. $\abs{\eta(x)}=0 \forall x \in \partial B_r(x_e)$. In case $x\in\mathbb{R}$, this yields the integrated area to be a triangle with height $h$, width $2r$ and area $\frac{1}{2}h2r\geq h\frac{h}{2L}=\frac{h^2}{2L}$ (Lemma S3).

\end{proof}

\textbf{Theorem 1.} Let $N$ be the sample size, and $a$ the treatment with the higher number of observations in $B_r(x_e)$, and $x\in \mathbb{R}$. Then the expected probability of Type S error in $B_r(x_e)$ has lower bound \\
$p(\text{``Type S error''}) > 2K^2 D^2 C^{N (P_{B_r(x_e)}^a - (1-p(a))\frac{h^2}{2L})}$.

Theorem 1 shows that, with fixed $r$, $N$ and $p(a)$, the larger the local imbalance ($h$) in $B_r(x_e)$, the higher Type S error rate in $B_r(x_e)$ is. 

\begin{proof}[Proof of Theorem 1]
The expected number of samples of group $a$ in $B_r(x_e)$ is $\E[n_a] = p(a)NP_{B_r(x_e)}^a$. The expected Type S error over all samples of size N from the true distribution is proportional to $\E[C^{(n_1 + n_0)}] \geq C^{(\E[n_1] + \E[n_0])}$ (Lemma S1 and Jensen's inequality).

From this and Lemma S4 it follows that the expected Type S error in $B_r(x_e)$ has lower bound
\begin{align*}
p&(\text{``Type S error''}) > 2K^2 D^2 C^{\E[n_a] + \E[n_{1-a}]} \\
& \geq 2K^2 D^2 C^{ \left(p(a)NP_{B_r(x_e)}^a + (1-p(a))N(P_{B_r(x_e)}^a - \frac{h^2}{2L}) \right)} \\
& \geq 2K^2 D^2 C^{N \left(P_{B_r(x_e)}^a - (1-p(a))\frac{h^2}{2L}) \right)}.
\end{align*}
\end{proof}

In higher dimension, the key difference is in the result of Lemma S4, affecting the term $\frac{h^2}{2L}$. Specifically, the integral $\int_{B_r(x_e)}\abs{\eta(x)}dx \geq M$, where $M$ depends on the dimensionality of $x$; in one dimension $M=\frac{h^2}{2L}$ as in Lemma S4.

Now, we have shown that imbalance increases locally the Type S error rate. Then the question remains whether the error rate increases globally as well, or do the local effects cancel out each other. We prove this in one-dimensional case, but we see no reason why the proof would not extend to higher dimensions as well. The following assumption and theorem give conditions under which imbalance increases the global Type S error rate.

\textbf{Assumption S5.} Assume the following balanced and imbalanced settings. In the balanced setting, let \\
$p^1(x)=p^0(x)=p(x)$, and $x\in \mathbb{R}$. Without loss of generality we assume that imbalance arises from a shift in $p^0(x)$, s.t. in the imbalanced setting $p^0(x) = p^1(x) - \eta(x)$, where $\eta(x) \in\mathbb{R}$, and $\int \eta(x) dx = 0$.

\textbf{Lemma S5.} In the imbalanced setting and under assumption S5, $p(x) = u p^1(x) + (1-u) p^0(x) = p^1(x) - (1-u)\eta(x)$, where $u:=p(a=1)$.

\textit{Proof.} By simply: $p(x) = u p^1(x) + (1-u) p^0(x) = u p^1(x) + (1-u) (p^1(x) - \eta(x)) = u p^1(x) + (1-u)p^1(x) - (1-u)\eta(x) = p^1(x) - (1-u)\eta(x)$.

\textbf{Lemma S6.} Given assumption S3, the maximum probability density at $x\in \mathcal{X}$ is $p_{\text{max}} \leq \sqrt{L}$.
 
\begin{proof}[Proof:]
By Assumption S3,
 \begin{align*}
     \abs{p^1(x)-p^1(x')} &\leq L\abs{x-x'} \text{,~~~~~and} \\
     \abs{p^0(x)-p^0(x')} &\leq L\abs{x-x'}  \text{,~~~~~and} \\
     p(x) = up^1(x) + &(1-u)p^0(x), \\
     \Rightarrow \abs{p(x)-p(x')} & \leq u L\abs{x-x'} + (1-u)L\abs{x-x'} \\
     &= L\abs{x-x'}.
 \end{align*}

Because $p(x)$ integrates to one, the highest possible density is achieved by first increasing $p(x)$ as quickly as possible to $p_{\text{max}}$, and then decreasing it back to zero; Otherwise some of the density would be spread to a wider range. Therefore, we get the maximum  $p_{\text{max}}$ by the sum of two triangles with height $p_{\text{max}}$ and width $\frac{p_{\text{max}}}{L}$:
\begin{align*}
    & 2*\frac{1}{2}p_{\text{max}} (\frac{p_{\text{max}}}{L}) \leq 1 \\
    \Leftrightarrow & p_{\text{max}} \leq \sqrt{L}.
\end{align*}
\end{proof}

The following theorem gives a sufficient condition for the increase of the expected global Type S error rate.

\textbf{Theorem 2}. Denote $P_{\eta \geq h}=\int_{\mathcal{X}}\mathbb{I}(\eta(x) \geq h)p_t(x)dx$, where $p_t(x)$ is the covariate distribution in the test set. Given Assumption S5, if $P_{\eta \geq h} > C^{N(1-u)h}$, then imbalance $\eta(x)$ increases the lower bound of the expected global Type S error rate in $\mathcal{X}$.

\begin{proof}[Proof of Theorem 2.]
We prove this in one-dimensional setting. The intuition is that since the error rate increases exponentially with decreasing number of samples, then in high-imbalance areas, where $\eta(x) \geq h$, the local increase in the error rate cannot be compensated elsewhere. We start by decomposing the lower bound to the bound without imbalance and a term that depends on imbalance. We then show that the imbalance-related term is greater than zero when $P_{\eta \geq h}$ is high enough, and therefore the imbalance increases the lower bound of the global Type S error rate.

In an infinitesimally small interval $dx$, the expected number of observations \textit{over all samples} of size N from the true distribution, is $\rho_1 dx= \E[n_1] = uNp^1(x)dx$ and \\
$\rho_0 dx= \E[n_0] = (1-u)Np^0(x)dx$.

Then, by Assumption S5, $\rho_1 + \rho_0 = uNp^1(x) + (1-u)N(p^1(x)-\eta(x)) = Np^1(x)-(1-u)N\eta(x) = N(p^1(x)-(1-u)\eta(x))= Np(x)$. (Last equation from Lemma S5).

Then the expected effect on the expected Type S error is proportional to $\E[C^{(n_1 + n_0)}] \geq C^{(\E[n_1] + \E[n_0])} = C^{(\rho_1 + \rho_0)}$ (Jensen's inequality and Lemma S1), and the expected error rate in $\mathcal{X}$ is
\begin{align*}
    \gamma &\geq 2K^2 D^2 \int_\mathcal{X} C^{(\rho_1 + \rho_0)}p_t(x)dx \\
    & = 2K^2 D^2 \int_\mathcal{X} C^{Np(x)}p_t(x)dx. \\
\end{align*}

Denote the expected error rate in the balanced setting as $\gamma_0 \geq 2K^2 D^2 \int_\mathcal{X} C^{Np^1(x)}p_t(x)dx := b_0$, which comes from the Assumption S5. Then the expected error rate in the imbalanced setting has a lower bound

\begin{align*}
    \gamma & \geq 2K^2 D^2 \int_\mathcal{X} C^{Np(x)}p_t(x)dx \\
    \intertext{which by Lemma S5 is}
    & = 2K^2 D^2 \int_\mathcal{X} C^{N(p^1(x) - (1-u)\eta(x))}p_t(x)dx \\
    & = 2K^2 D^2 \int_\mathcal{X} \Big( C^{Np^1(x)} C^{- N(1-u)\eta(x)} \\
    & \text{~~~~~~~~~~~~~~~~~~~~~~~~~~~~} - C^{Np^1(x)} + C^{Np^1(x)}\Big)p_t(x)dx \\
    & = 2K^2 D^2 \int_\mathcal{X}  C^{Np^1(x)} \left(C^{- N(1-u)\eta(x)}  - 1\right)p_t(x)dx \\
    & \text{~~~~~~~~~~~~~~~~~~~~~~~~~~~~} + 2K^2 D^2\int_\mathcal{X}C^{Np^1(x)}p_t(x)dx \\
    & = 2K^2 D^2 \int_\mathcal{X}  C^{Np^1(x)} \left(C^{- N(1-u)\eta(x)}  - 1\right)p_t(x)dx + b_0 \\~
    & \geq 2K^2 D^2 \int_\mathcal{X}  C^{Np_{\text{max}}} \left(C^{- N(1-u)\eta(x)}  - 1\right)p_t(x)dx + b_0, \\
    \intertext{and by Lemma S6}
    & \geq 2K^2 D^2 C^{N\sqrt{L}}\left( \int_\mathcal{X} C^{- N(1-u)\eta(x)}p_t(x)dx  - 1\right) + b_0.
\end{align*}

Since $b_0$ is the lower bound in the balanced setting, the lower bound of the expected Type S error rate increases with increasing imbalance, if $\int_\mathcal{X} C^{- N(1-u)\eta(x)}p_t(x)dx  > 1$.

Next, we consider when does this condition hold. Denote the set where $\eta(x)\geq0$ as $\mathcal{X}^+$, and similarly $\mathcal{X}^-$ the set where $\eta(x) < 0$. Then
\begin{align*}
    & \int_\mathcal{X} C^{- N(1-u)\eta(x)}p_t(x)dx \\ 
    & = \int_{\mathcal{X}^+} C^{- N(1-u)\abs{\eta(x)}}p_t(x)dx \\
    & + \int_{\mathcal{X}^-} C^{N(1-u)\abs{\eta(x)}}p_t(x)dx \\
    & \geq \int_{\mathcal{X}^+\backslash \mathcal{X}_{\eta \geq h}} C^{- N(1-u)\abs{\eta(x)}}p_t(x)dx \\
    &+ \int_{\mathcal{X}_{\eta \geq h}} C^{- N(1-u)h}p_t(x)dx
    + \int_{\mathcal{X}^-} C^{N(1-u)\eta_{\text{max}}}p_t(x)dx \\
    & \geq \int_{\mathcal{X}^+\backslash \mathcal{X}_{\eta \geq h}} p_t(x)dx \text{~}+\text{~}  C^{- N(1-u)h} \int_{\mathcal{X}_{\eta \geq h}} p_t(x)dx \\
    &+ C^{N(1-u)\eta_{\text{max}}} \int_{\mathcal{X}^-}  p_t(x)dx \\
    & = P_{0 \leq \eta < h} \text{~}+\text{~} C^{- N(1-u)h}P_{\eta \geq h} \text{~}+\text{~} C^{N(1-u)\eta_{\text{max}}}P_{\eta<0} \\
    & \geq C^{- N(1-u)h}P_{\eta \geq h} \\
    & > 1 \text{,~~~~ if~~} P_{\eta \geq h} > C^{N(1-u)h}.
\end{align*}

Here $\eta_{\text{max}}$ is the maximum difference between the the distributions $p^1(x)$ and $p^0(x)$, and \\
$\mathcal{X}_{\eta \geq h} = \{x\in \mathcal{X} \mid \eta(x) \geq h \}$, $h>0$.
\end{proof}

\section{Details of the Implementations}
\subsection{The Observed and Estimated Type S Error Rate in Imbalanced Data}
\label{sec:supp_error_rates}
Data is generated from 
\begin{align*}
    x & \sim N(0,1) \\
    a & \sim \bernoullipdf(\theta_x) \\
    b_0, b_1 & \sim N(0,0.5) \\
    y \mid x,a & \sim N(f(x)+(\beta_0 + \beta_1x)a,\sigma_0^2), \text{ and} \\
    f(x) &= 2\left(\frac{1}{1+e^{-x+b}}-0.5\right),
\end{align*}
where imbalance is generated by setting $\theta_x = e_x$ for $x\leq0$ and $\theta_x=1-e_x$ for $x>0$. Here $e_x=p(a=1\mid x)$ is the propensity score. Technical details: The shape of $f(x)\in (-1,1)$ is chosen to be half of a sigmoid within range of $1\sigma$ from $\bar{x}$, so as to either have a saturating effect or an increasingly increasing effect (defined by the sign of $b\in\{-1,1\}$, $b \sim \text{uniform}$).

The outcome model generation ($b_0,b_1$ and $b$) is repeated 200 times, and for each outcome model we generate 6 training sets. The training data generation differs in the propensity scores $e_x\in\{0.0,0.1,...,0.5\}$, resulting in different levels of imbalance in the training data sets. The size of the test set is $500$.

We measure imbalance using the Maximum Mean Discrepancy (MDD)~\cite{gretton2012kernel}, with Gaussian kernel and length-scale 0.8. We model the potential outcomes using two independent Gaussian Processes with squared exponential kernel.

\subsection{Simulated Example}
\paragraph{Synthetic data.} The outcome $y\in \{0,1\}$ is Bernoulli distributed with parameter $\theta_{x,a}$, given a one-dimensional covariate $x \in \mathbb{R}$ and treatment $a \in \{0,1\}$. The data is generated from a logistic regression model with interaction between $a$ and 3 radial basis functions (RBF) $\phi(x)$. The data is generated as follows:
\begin{align*}
    x &\sim \text{uniform}(-4.5,4.5) \\
    a \mid x &= 1 \text{~~if~~} x <-1.5,\text{~~} 0 \text{~~else}\\
    y \mid x,a &\sim \bernoullipdf(\theta_{x,a}),\\
\end{align*}
where $\theta_{x,a} = \text{logit}^{-1}(\mathbf{w}_0^\top\phi(x) + \mathbf{w}_1^\top \phi(x)a)$, and RBF centers are at $-3,0,3$, have lenght-scale $1$, and $\mathbf{w}_0=[0.5~ 1.5~ 1.5]^\top$, $\mathbf{w}_1=[1~ -1~ -3.0]^\top$.

Training sample size is 30. The 9 test points are set with equal distance to each other in the range of $x$. Data generation is repeated 100 times, and the reported values are the mean and the 95\% bootstrap confidence intervals.

\paragraph{Model and learning.} We model the data with a logistic regression model $p(y\mid x, a)\sim \bernoullipdf(\theta_{x,a})$, where $\theta_{x,a}$ has the same form as in the data generation process. The model is fit using a probabilistic programming language Stan~\citep{pystan21600, stan2017}. We assume that the RBF centers and length-scale are known ($-3,0,3$, and lenght-scale $1$), so that only $\mathbf{w}_0$ and $\mathbf{w}_1$ need to be learned.

\subsection{Gaussian Process Model with Direct Feedback}

For modeling the Gaussian process with direct feedback on patient response, a Gaussian process prior with squared exponential covariance function and Gaussian likelihood was used. Responses for different treatments were modeled with independent models. We use Gamma distribution with shape 1.5 and rate 3.0 as prior for lengthscale, variance and noise. The models were implemented with GPy-framework \footnote{GPy v. 1.9.2 \url{https://github.com/SheffieldML/GPy}}. Since the observed data and the counterfactual feedback were obtained from different sources, both were assumed to have separate noise priors. Since different attributes have very different effect on the response, the covariance function used different lengthscale parameters for different dimensions. Hyper-parameters were estimated by maximizing the marginal likelihood.

\section{Additional results}

\subsection{The Observed and Estimated Type S Error Rate in Imbalanced Data}
\begin{figure}
    \centering
    \includegraphics[width=0.78\linewidth]{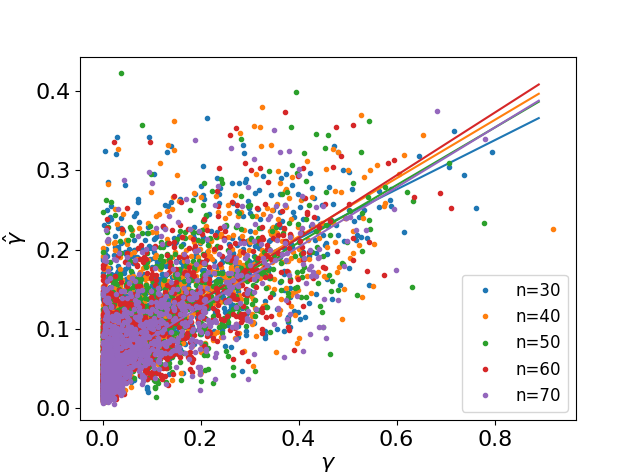}
    \caption{The observed ($\gamma$) and estimated ($\hat{\gamma}$) Type S error rate in 1200 data sets. Low estimated error indicates low observed error. Solid lines show the regression line for each sample size.}
    \label{fig:true_vs_estm}
\end{figure}

The observed and estimated Type S error rates from the experiment described in Section~\ref{sec:supp_error_rates} (Section 5.1 in the paper) are shown in Figure~\ref{fig:true_vs_estm}. The results show that low estimated error indicates low observed error.

\subsection{K-Nearest Neighbor Approximation of D-M aware}
\begin{figure}
    \centering
    \includegraphics[width=0.4\textwidth]{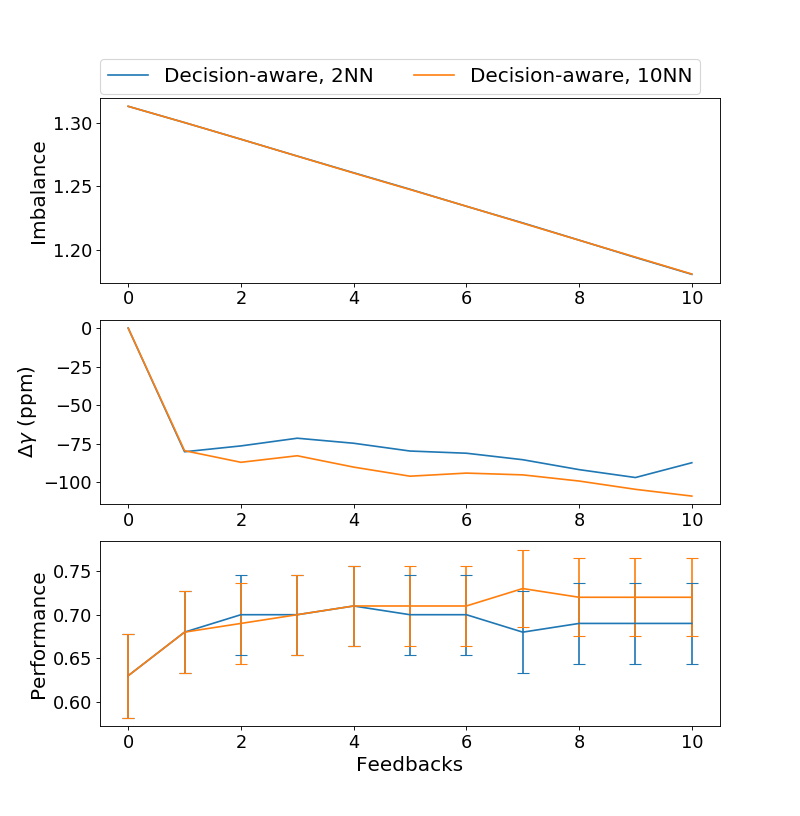}
    \caption{Comparison of the D-M aware active learning using k=2 (Decision-aware 2NN) and k=10 (Decision-aware 10NN) nearest neighbor approximations shows that the performance degrades if the number of neighbors k is too low.}
    \label{fig:supp_knn}
\end{figure}

We additionally tested the idea to approximate full D-M aware by only computing the expected minimization of Type S error rate for the k nearest neighbors of the test unit. Important here is that we use the \textit{model's distance measure}, which in the case of GPs is the optimized kernel (with Automatic Relevance Determination ARD). Our preliminary results (Fig.~\ref{fig:supp_knn}) show that selecting too few neighbors will impair the performance of active learning.

\end{document}